\documentclass[conference]{IEEEtran}

\usepackage{graphicx}
\usepackage{amsmath}
\usepackage{cite}

\makeatletter

\newcommand{\Rmnum}[1]{\expandafter\@slowromancap\romannumeral #1@}
\makeatother

\begin{document}
%
\title{Review of Applications of Generalized Regression Neural Networks in Identification and Control of Dynamic Systems}

\author{\IEEEauthorblockN{{Ahmad Jobran Al-Mahasneh, Sreenatha G. Anavatti and Matthew A. Garratt}}
\IEEEauthorblockA{School of Engineering and Information Technology \\ The University of New South Wales at the Australian Defense Force Academey\\ Canberra, ACT 2612, Australia\\Emails: ahmad.al-mahasneh@student.adfa.edu.au; a.sreenatha@adfa.edu.au; M.Garratt@adfa.edu.au}}

\maketitle

\begin{abstract}
This paper depicts a brief revision of Generalized Regression Neural Networks (GRNN) applications in system identification and control of dynamic systems. In addition, a comparison study between the performance of back-propagation neural networks and GRNN is presented for system identification problems. The results of the comparison confirms that, GRNN has shorter training time and higher accuracy than the counterpart back-propagation neural networks. 
\end{abstract}


\IEEEpeerreviewmaketitle

\section{Introduction}
Artificial Intelligence (AI) has a significant impact on the current research trends due its numerous applications in different aspects of the life.
Artificial Neural Networks(ANNs)  are one of the major parts of AI. ANNs have different applications including regression and approximation, forecasting and prediction, classification, pattern recognition and more.
ANNs are useful since they can learn from the data and they have global approximation abilities. A feed-forward neural network with at least single hidden layer and sufficient number of hidden neurons can approximate any arbitrary continuous function under certain conditions~\cite{hornik1989multilayer}. 
ANNs have two main types: the Feed Forward ANNs (FFANNs) in which the input will only flow to the output layer in the forward direction and the Recurrent ANNs (RANNs) in which  data flow can be in any direction.
Generalized Regression Neural Networks (GRNN)~\cite{specht1991general} are single-pass associative memory feed-forward type Artificial Neural Networks (ANNs) and uses normalized Gaussian kernels in the hidden layer as activation functions.

\begin{figure}[b]

\centering
\includegraphics[width=3.5in]{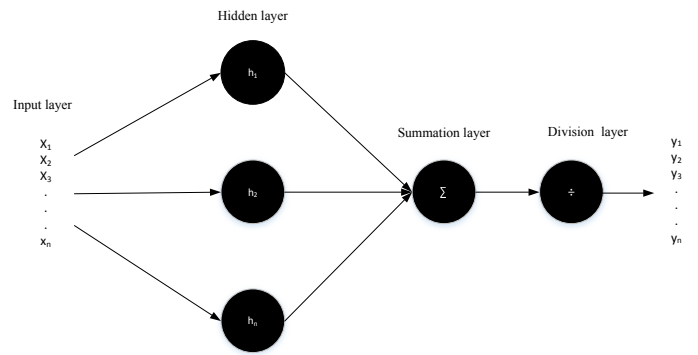}%
\caption{GRNN structure}
	\label{fig:grnn}
\end{figure}

GRNN is made of input, hidden, summation , division layer and output layers as shown in ~\figurename{~\ref{fig:grnn}}.

When GRNN is trained, it memorizes every unique pattern. This is the reason why it is single-pass network and does not require any back-propagation algorithm.

After training GRNN with adequate training patterns, it will be able to generalize for new inputs. The output of GRNN can be calculated using ~\eqref{eq2:grnn1} and ~\eqref{eq2:grnn2}.

\begin{equation}
D_{i}=(X-X_{i})^{T}(X-X_{i})
\label{eq2:grnn1}
\end{equation}
\begin{equation}
\hat{Y}=\frac{\sum_{i=1}^{N} Y e^{(-D_{i}/2\sigma^{2})}}{\sum_{i=1}^{N} e^{(-D_{i}/2\sigma^{2})}}
\label{eq2:grnn2}
\end{equation}

\noindent where $D_{i}$ is the Euclidean distance between the input $X_{i}$ and the training sample input $X$, $Y$ is the training sample output, $\sigma$ is the smoothing parameter of GRNN.

GRNN advantages include its quick training approach and its accuracy. On the other hand, one of the disadvantage of GRNN is the growth of the hidden layer size. However, this issue can be solved by implementing a special algorithm which reduces the growth of the hidden layer by storing only the most relevant patterns~\cite{Al-Mahasneh2017grnnfwmavhawaii}.

In this paper a brief review of the applications of GRNN in modeling and identification and control is provided.  A comparison also is conducted between the GRNN and back-propagation neural networks for various benchmarking problems. The rest of the paper is structured as follows: section \Rmnum{2} provides a review of the applications of GRNN in system identification and modeling, section \Rmnum{3} offers a review of the GRNN applications in control systems, section \Rmnum{4} contains the results of GRNN and back-propagation neural networks comparisons and their discussions and finally section \Rmnum{5} contains  conclusions of the research.

\section{Applications of GRNN in modeling and system identification}
\begin{figure}[h]
	\centering
	\includegraphics[width=3.5in]{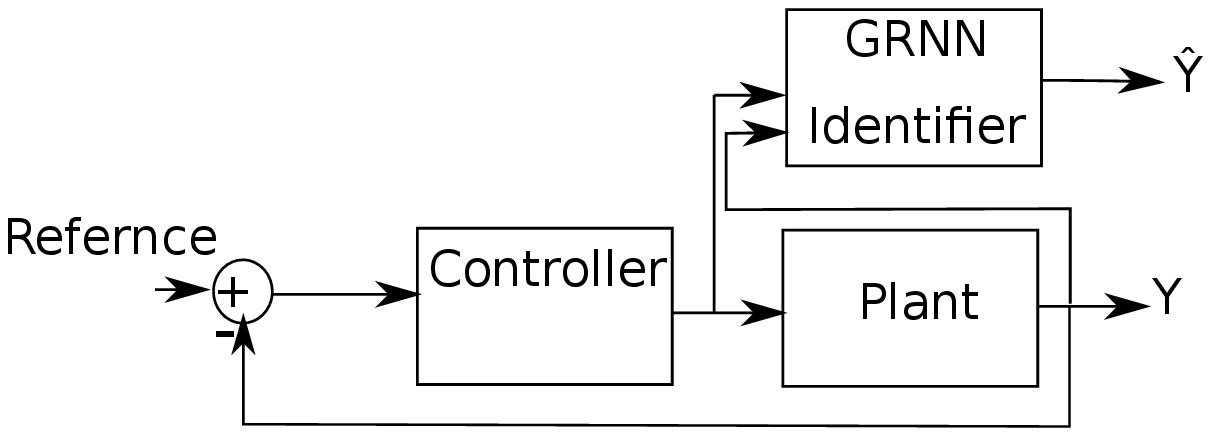}%

	\caption{GRNN structure}
		\label{fig:grnn_sysid}
\end{figure}

GRNN can be used as identifier for a given plant dynamics or a process. To use GRNN as identifier it should be given the plant/process inputs and outputs and then it will predict the plant/process output. GRNN identifier is shown in ~\figurename{\ref{fig:grnn_sysid}}, where $Y$ represent the actual plant output and $\hat{Y}$ is the predicted plant output.

Since GRNN is a regression-based neural network, it is widely used for approximation, fitting, prediction and regression problems including process modeling and monitoring~\cite{kulkarni2004modeling}, modeling of dynamic plants~\cite{seng2002adaptive}, aerodynamic forces prediction~\cite{yao2012three}, prediction of voltage variations in power plants~\cite{das2009use},prediction of performance and exhaust emissions in internal combustion engine~\cite{bendu2016application}, solar photo voltaic power forecasts~\cite{alhakeem2015new}, traffic accidents prediction~\cite{liu2007traffic}, nonlinear radio frequency systems modeling~\cite{li2011researches}, predictive modeling of non-linear systems~\cite{song2005predictive}, modeling of microwave transistors~\cite{gunecs2017cost} and Identification of a quadcopter dynamics~\cite{Al-Mahasneh2017quaddoha}.

\section{Applications of GRNN in control systems}
\begin{figure}[h]
	\centering
	\includegraphics[width=3.5in]{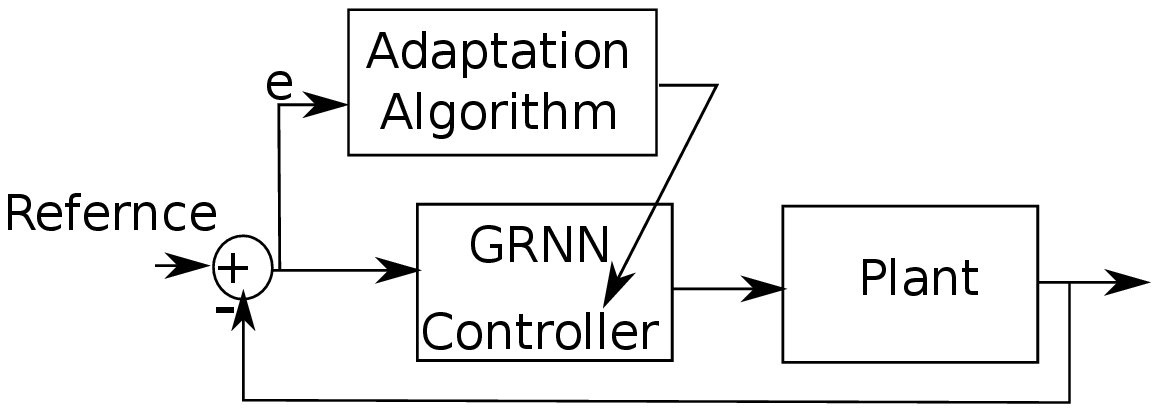}%
	\caption{GRNN structure}
		\label{fig:grnn_controller}
\end{figure}

In addition to its application in system identification and modeling, GRNN can be used as intelligent adaptive controller. GRNN can be used in control systems either off-line or on-line. If it is used off-line, it needs to be trained to work as inverse dynamics and then it will be deployed into the on-line system. The second approach is using GRNN controller on-line from the beginning so it will adapt to the error signals on-line. In the second case an adaptation algorithm is required to adjust GRNN parameters. The second approach is shown in ~\figurename{\ref{fig:grnn_controller}}

Some of the applications of GRNN in control systems include dead-zone estimation and compensation in motion control of a traveling wave ultrasonic motor~\cite{chen2009motion}, fault diagnosis of power system~\cite{qing2005adaptive}, intelligent battery charger~\cite{petchjatuporn2006implementation}, microgrid hybrid power systems control~\cite{ou2014dynamic}, bipedal standing stabilization~\cite{ghorbani2007nearly}, air conditioning control~\cite{ben2002energy}, wind generation system~\cite{hong2014optimal}, helicopter motion control~\cite{amaral2003helicopter}, active vibration control~\cite{madkour2007intelligent}, active noise cancellation~\cite{salmasi2011evaluating}, rat-like robot control~\cite{sun2011automatic}, pipe climbing robot control~\cite{fan2012implementation}, tracking-control for an optomechatronical Image derotator~\cite{altmann2016identification}, tracking in marine navigational radars~\cite{stateczny2008determining}, factory monitoring~\cite{su2011comparison}, and flapping wing micro aerial vehicle control~\cite{Al-Mahasneh2017grnnfwmavhawaii}.

\section{GRNN vs Back-propagation ANN}
In this section a comparative study is conducted between the GRNN performance and  a Back-Propagation (BP) FFANNs in the training stage.

 The deployed datasets  are provided by Matlab Mathworks~\cite{matlab} as a benchmarking data for ANNs performance measuring. Here are descriptions of the used datasets:
 
 	\begin{itemize}
 		\item The simple fit dataset is a 1-input/1-output fitting dataset and it contains 94 observations.
 		\item Abalone shell rings dataset is 8-input/1-output fitting dataset and it contains 4177 observations. The ANN should estimate the number of shell rings in abalone based on the 8-inputs.
 		\item Building energy dataset is 14-input/3-output fitting dataset and it contains 4208 observations. The ANN should approximate the energy use in a building based on the given 4-inputs.
 		\item Cholesterol dataset is 21-input/264-output fitting dataset and it contains 4208 observations. The ANN should estimate the Cholesterol levels in the body based on the 21-inputs.
 		\item Engine dataset is 2-input/2-output fitting dataset and it contains 1199 observations. The ANN should estimate the engines torque and emissions based on the given 2-inputs.
 		\item Cancer dataset is 4-input/3-output classification dataset and it contains 150 observations. The ANN should classify the type of cancer based on the 4-inputs.
 		\item Thyroid dataset is 21-input/3-output clustering dataset and it contains 7200 observations. The ANN should classify the patient as normal, hyperfunction or subnormal functioning based on the given 21-inputs.
 	\end{itemize}

\begin{table*}[h]
	\centering
	\caption{GRNN Vs BP}
	\begin{tabular}{|c|c|c|c|c|}
		\hline
		Dataset&	GRNN training error (MSE)& GRNN training time(sec)&BP training error(MSE)&BP training time(sec)\\
		\hline
		Simple fitting dataset&4.44E-18&0.479&9.51E-5&0.516\\
		\hline
		Abalone shell rings dataset&0.0828&0.493&4.191&0.566\\
		\hline
		Building energy dataset&2.561E-6&0.488&0.0022&1.284\\
		\hline
		Cholesterol dataset&5.560E-5&0.515&3.670E+2&0.593\\
		\hline
		Engine behavior dataset&2.670E-10&0.494&2.420E+3&0.671\\
		\hline
		Breast cancer dataset&5.980E-29&0.489&0.023&0.604\\
		\hline
		Iris flower dataset&1.360E-21&0.511&0.009&0.533\\
		\hline 
		Thyroid function dataset&1.920E-4&0.490&0.007&4.995\\
		\hline
	\end{tabular}
		\label{table1:mse_training_time}	
\end{table*}

The Mean Square Error (MSE) and the training time are recorded for both of BP ANN and GRNN for different datasets and shown in ~\tablename{~\ref{table1:mse_training_time}}.

MSE is a prominent performance measure in ANNs. It can be calculated using the following equation:

\begin{equation}
MSE=\frac{1}{N}\sum_{i=1}^{N}(Y(n)-T(n))^{2}
\end{equation}
\noindent where $N$ is the size of the output vector, $Y$ is a vector of the output of the neural network, $T$ is a vector of the target values.

Based on the results in \tablename{~\ref{table1:mse_training_time}}, GRNN has always higher training accuracy than BP ANN and less training time. In fact, for Cholesterol and engine data set the training error for BP is very high while on the other hand for GRNN the error is very low.

\begin{figure}[t]
	\centering
	\includegraphics[width=3.5in]{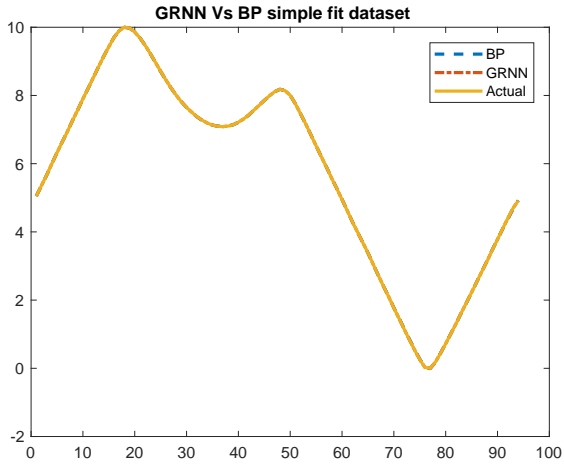}%
	\caption{GRNN Vs BP simple fit dataset}
		\label{fig:simplefit_grnn_vs_bp}
\end{figure}

\begin{figure}[t]
	\centering
	\includegraphics[width=3.5in]{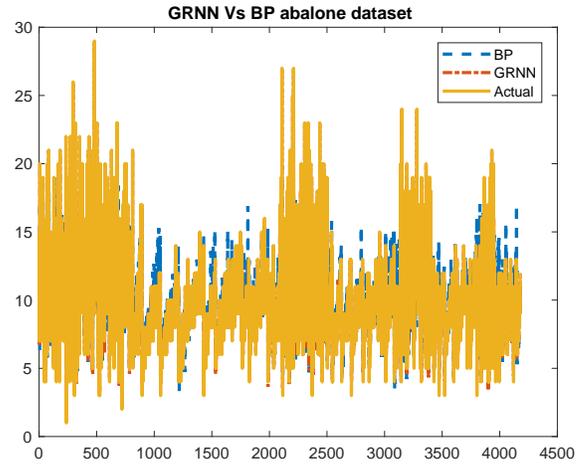}%
	\caption{GRNN Vs BP abalone dataset}
		\label{fig:abalone_grnn_vs_bp}
\end{figure}

\begin{figure}[t]
	\centering
	\includegraphics[width=3.5in]{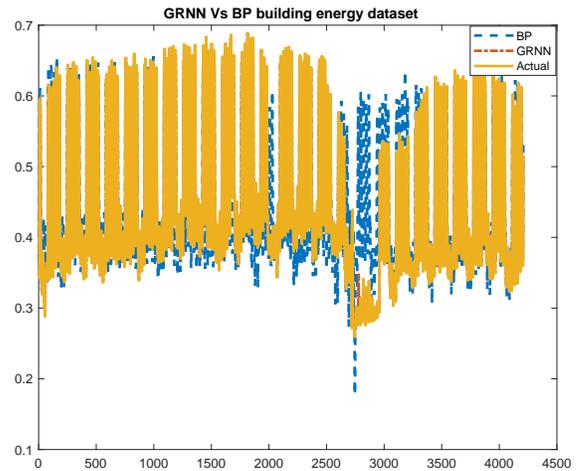}%

	\caption{GRNN Vs BP building energy dataset}
		\label{fig:building_energy_grnn_vs_bp}
\end{figure}

\begin{figure}[t]
	\centering
	\includegraphics[width=3.5in]{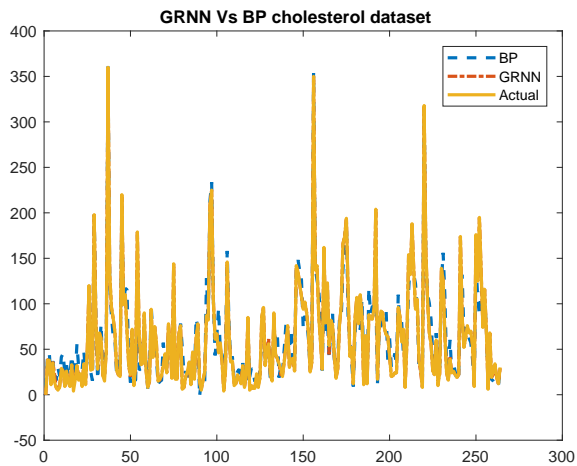}%
	\caption{GRNN Vs BP cholesterol dataset}
		\label{fig:cho_grnn_vs_bp}
\end{figure}

\begin{figure}
	\centering
	\includegraphics[width=3.5in]{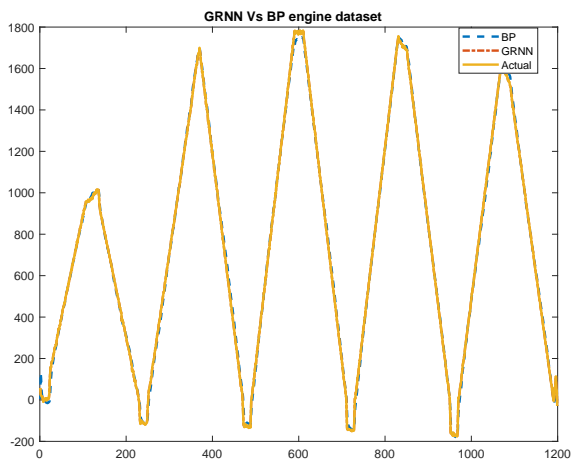}%
	\caption{GRNN Vs BP engine dataset}
		\label{fig:engine_grnn_vs_bpp}
\end{figure}

\begin{figure}[t]
	\centering
	\includegraphics[width=3.5in]{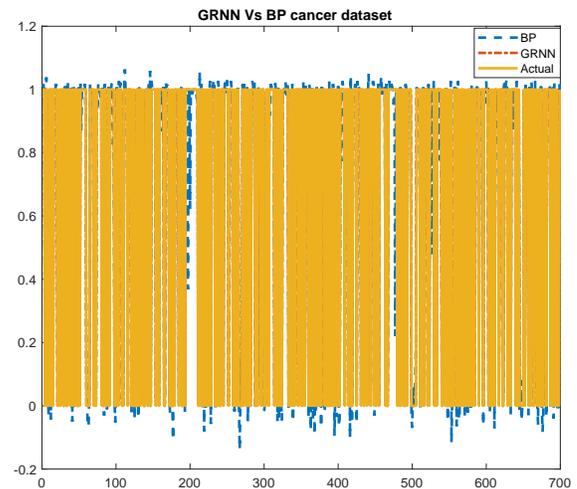}%
	\caption{GRNN Vs BP cancer dataset}
		\label{fig:cancer_grnn_vs_bp}
\end{figure}

\begin{figure}
	\centering
	\includegraphics[width=3.5in]{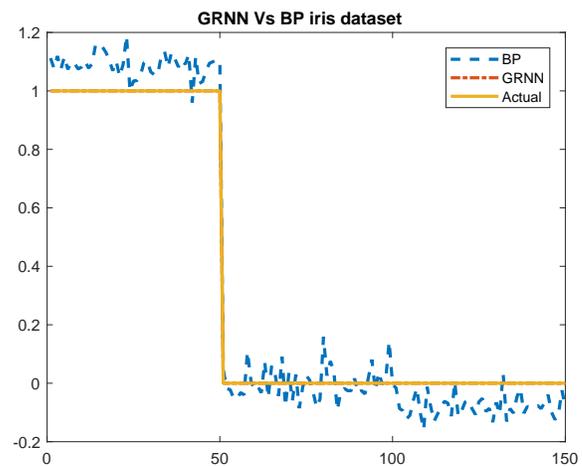}%
	\caption{GRNN Vs BP iris dataset}
		\label{fig:iris_grnn_vs_bp}
\end{figure}

\begin{figure}[t]
	\centering
	\includegraphics[width=3.5in]{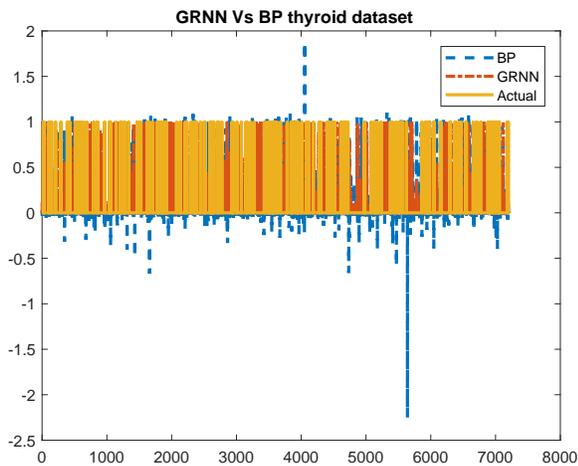}%
	\caption{GRNN Vs BP thyroid dataset}
		\label{fig:thyroid_grnn_vs_bp}
\end{figure}

In the second set of results, the output from every dataset is used to compare bewteen GRNN and BP as shown in ~\figurename{~\ref{fig:simplefit_grnn_vs_bp}}, ~\figurename{~\ref{fig:abalone_grnn_vs_bp}}, ~\figurename{~\ref{fig:building_energy_grnn_vs_bp}}, ~\figurename{~\ref{fig:cho_grnn_vs_bp}}, ~\figurename{~\ref{fig:engine_grnn_vs_bpp}}, ~\figurename{~\ref{fig:cancer_grnn_vs_bp}}, ~\figurename{~\ref{fig:iris_grnn_vs_bp}}, and ~\figurename{~\ref{fig:thyroid_grnn_vs_bp}}.

\begin{figure}[t]
	\centering
	\includegraphics[width=3.5in]{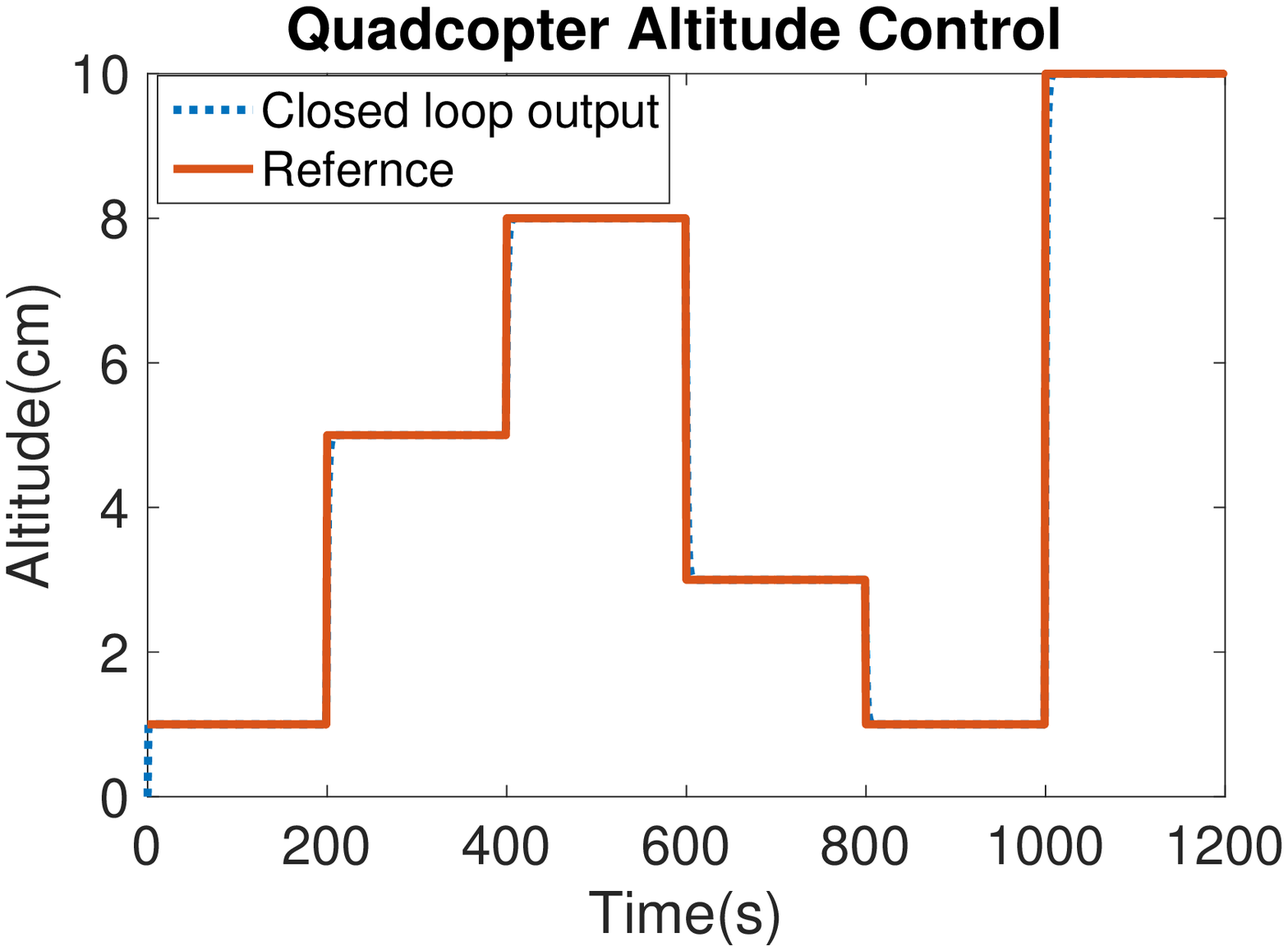}%
	\caption{GRNN Quadcopter  altitude tracking}
	\label{fig:grnn_alt_quadcopter_controller}
\end{figure}

In the last part of the results, GRNN is used to control the altitude of a quadcopter model in the simulation using Matlab. The altitude tracking of the quadcopter is shown in~\figurename{\ref{fig:grnn_alt_quadcopter_controller}}. The GRNN controller is accurately follow the set point reference and the learning is quick.



\section{Conclusion}
GRNN provides accurate and quick solution to regression, approximation, classification and fitting problems.\\
GRNN can be used in system identification of dynamic systems as well as control of dynamic systems.\\

GRNN outperforms BP ANNs in the accuracy and training time; however, GRNN has some limitations such as the growth of the hidden layer.\\

When training GRNN with a large dataset it is essential to reduce the data dimensionality using any of the data reduction techniques such as clustering or distance based algorithms.

\bibliographystyle{IEEEtran}
\bibliography{IEEEabrv,bib}

\end{document}